\documentclass[letterpaper, 10 pt, conference]{ieeeconf}

\IEEEoverridecommandlockouts
\usepackage{cite}
\usepackage{amsmath,amssymb,amsfonts}
\usepackage{algorithmic}
\usepackage{graphicx}
\usepackage{textcomp}
\usepackage{xcolor}
\usepackage{booktabs}
\usepackage{bm}
\usepackage{caption}
\def\BibTeX{{\rm B\kern-.05em{\sc i\kern-.025em b}\kern-.08em
    T\kern-.1667em\lower.7ex\hbox{E}\kern-.125emX}}
\usepackage{hyperref}

\newif\ifshowdiff
\showdifffalse    

\usepackage[normalem]{ulem}
\usepackage{color}

\title{SalsaAgent: A multimodal embodied language model for interactive dance generation}

\author
{
    Payam Jome Yazdian, Zoe Stanley, and Angelica Lim
    \thanks{All authors are associated with Simon Fraser University, Burnaby, Canada {\tt\small \{payam\_jome-yazdian, zoe\_stanley, angelica\}@sfu.ca}}
}

\begin{document}

\maketitle
\thispagestyle{empty}
\pagestyle{empty}
\begin{abstract}
\ifshowdiff
    \input{sections_diff/0_Abstract_diff}
\else
Embodied interaction with humanoids depends on bidirectional nonverbal reactivity, coordination, and synchrony to convey cues and move with a partner. For socially interactive embodied agents, reactive motion generation requires expressive full-body motion that remains contextually appropriate while maintaining spatial and temporal synchrony. 
We present SalsaAgent, a language model that generates expressive, full-body salsa follower motions in reaction to a human leader and music.
We formulate partner interaction as nonverbal token passing, extending the vocabulary of a large language model (LLM) to process discrete motion tokens, pairwise relation tokens, and audio tokens. Our method introduces full-body and pairwise-relation tokenizers, aligns language and motion tokens with automatically derived text descriptions of skeleton dynamics, and applies a two-stage token-to-diffusion pipeline. Subjective and objective evaluations show improved motion quality, two-person spatial coordination, and music and partner coordination relative to prior baselines.
\\Project page: \href{https://pjyazdian.github.io/Salsa-Agent}{https://pjyazdian.github.io/Salsa-Agent}

\fi
\end{abstract}


\ifshowdiff
    \input{sections_diff/1_Introduction_diff}
    \input{sections_diff/2_RelatedWork_diff}
    \input{sections_diff/3_Method_diff}
    \input{sections_diff/4_Experiment_diff}
    \input{sections_diff/5_Results_Discussion_diff}
    \input{sections_diff/5_Conclusion_diff}
\else
    \section{Introduction}

Embodied interaction beyond speech depends on nonverbal reactivity, coordination, and synchrony as fundamental components of socially interactive embodied agents\cite{bartneck2024human}.
Social dance interactions~\cite{hanna1987dance, 10.1145/3323335}, such as salsa~\cite{McMains02012016}, offer a compelling example of interactive nonverbal communication, where coordinated movement under an explicit lead--follow dynamic requires the follower to interpret the leader's motion and respond with appropriate movements.
Accordingly, we study reactive dance generation: given a leader's motion and music, generate the follower's full-body motion with coordinated, contextually appropriate response. This can provide a kinematic motion reference in the form of skeleton-level trajectories as an upstream motion-generation component for retargeting to socially interactive embodied agents~\cite{yang2025omniretarget}.

Current methods for reactive dance generation only partially address the challenges in this task. 
Existing methods~\cite{siyao2024duolando, liang2024intergen, li2024interdance} fail to effectively model close interaction dynamics, often producing followers that face the wrong direction, lose spatial connection with their partner and exhibit foot-skating artifacts due to the inherent limitations of their motion representations.
A promising direction is fine-tuning large language models (LLMs) to interact using motions instead of text, but current dance datasets lack the fine-grained language annotations needed to train them in nonverbal bodily communication~\cite{hanna1987dance}.

We introduce SalsaAgent, a multimodal embodied LLM that formulates  human-human interaction as an embodied language and extends the language model vocabulary to incorporate motion, pairwise relational tokens, and audio as context~\cite{jiang2024motiongpt, yu2025socialgen}. To fine-tune a large language model to process motion tokens, we provide large-scale text--motion supervision by grounding a duet dance dataset~\cite{burkanova2025compas3d} with $300$K strided short samples annotated with fine-grained body-part movement descriptions using an automatic rule-based algorithm~\cite{yazdian2023motionscript}.
We devise two new VQ-VAE codebooks: one for single-person full body motion, and another for pairwise relations representing partner geometry.
Moreover, we train a motion diffusion prior model as a refinement stage in shared joint-trajectory space to further improve inter-dancer positioning, timing, and contact detail.
Objective and subjective evaluation shows our method improves on existing leader-to-follower methods in realism, partner coordination, and beat alignment.

Our contributions are as follows:

\begin{itemize}
    \item We introduce SalsaAgent, a multimodal embodied LLM  that formulates interaction as nonverbal motion token passing, improving follower dance motion generation over prior methods.
    \item We propose new VQ-VAEs for single-person full-body motion and pairwise relations.
    \item We show that an important factor is the automatic creation of fine-grained body-part movement descriptions to fine-tune the base language model.
    \item We show that a diffusion-based refinement stage in shared joint-trajectory space further improves inter-dancer positioning, timing, and contact details.
\end{itemize}

    \section{Related Work}
We summarize prior work on multimodal language models for motion, human motion datasets, human motion generation and understanding, and human--human interaction modeling.

\textbf{Human Motion Generation and Understanding.}
Human motion understanding and generation are fundamental challenges in animation and robotics~\cite{bartneck2024human, zhang2024react}. Humanoid systems often retarget kinematic motion references for whole-body locomotion, manipulation, and scene interaction, making motion generation a promising step toward socially interactive embodied agents\cite{yang2025omniretarget}.
Recent work has investigated motion generation conditioned on a range of inputs, including \emph{natural language}~\cite{guo2022cvpr_diverse, zhang2023generating}, \emph{audio and speech}~\cite{dabral2022mofusion, zhou2023ude}, \emph{music} (including dance)~\cite{siyao2022bailando, tseng2022edge, le2023music, le2023controllable}, 
and \emph{structured signals} such as action labels~\cite{petrovich21actor, guo2020action2motion}. 
Methods such as diffusion-based motion synthesis~\cite{tevet2022human, shafir2023human, nichol2021improved} and vector-quantized motion representation~\cite{van2017neural, zhang2023generating} with masked prediction~\cite{guo2023momask} are commonly used for single-person motion generation. 
Existing multi-person dance generation methods, such as group choreography~\cite{le2023music, le2023controllable} and duet dance approaches~\cite{siyao2024duolando, li2024interdance}, have shown promising results. However, limited motion representations prevent these methods from effectively capturing close interactions between two dancers.

\textbf{Human-Human Interaction Modeling.}
Human interaction presents several challenges, including semantic consistency, interpersonal coordination, and local interaction~\cite{sui2026survey}.
Prior work in human-human interaction modeling includes \emph{reaction synthesis} that generates reactive motion conditioned on observed motion~\cite{ liu2023interactive, ghosh2024remos, xu2024regennet} and \emph{joint generation} that generates multiple motions simultaneously~\cite{liang2024intergen, javed2024intermask, fan2024freemotion}.
Discrete-token methods such as Duolando~\cite{siyao2024duolando} and InterMask~\cite{javed2024intermask} quantize motion into compact tokens, enabling high-level motion representations but losing fine-grained details due to discretization and mode averaging.
In contrast, diffusion-based methods such as InterDance~\cite{li2024interdance}, ReMoS~\cite{ghosh2024remos}, and ReGenNet~\cite{xu2024regennet} operate in continuous space, achieving realistic motion generation while limiting long-range semantic dependencies. Overall, a gap persists in generating fine-grained motion that is also contextually appropriate over long sequences.

\textbf{Multimodal Language Models.}
Large language models (LLMs), such as LLaMA~\cite{touvron2023llama}, Vicuna~\cite{vicuna2023}, and Gemma~2~\cite{gemma2024report}, have shown strong performance in text understanding and generation~\cite{brown2020language}.
These models are increasingly extended with images~\cite{girdhar2023imagebind, li2022blip}, video~\cite{zhang2023video, wan2025}, audio~\cite{deshmukh2023pengi, deng2024musilingo}, and other modalities~\cite{han2023onellm, wu2023nextgpt}.
For motion-language modeling, MotionGPT~\cite{jiang2024motiongpt, zhang2024motiongpt} represents motion as a sequence of discrete tokens to align with language, Motion-Agent~\cite{wu2024motion} introduces conversational control, and M$^3$GPT~\cite{luo2024m3gpt} unifies multi-task motion comprehension and generation. Furthermore, MotionScript~\cite{yazdian2023motionscript} and FineMoGen~\cite{zhang2024finemogen} introduce fine-grained alignment between language and motion.
Recent efforts extend motion-language modeling to multi-human social interaction. Social Agent~\cite{zhang2025social}, and SOLAMI~\cite{jiang2025solami} focus on dyadic interaction settings, while SocialGen~\cite{yu2025socialgen} scales to multi-human scenarios within an LLM framework. 
In summary, existing work targets conversational or generic social scenarios, where \emph{speech or language} is used to condition interactions with LLMs for motion generation.
In contrast, LLM-based methods for reactive motion generation that communicate through \emph{nonverbal} motion tokens remain relatively underexplored. Furthermore, discrete token decoding often loses the fine-grained detail required for close partner interaction.

\textbf{Human Motion Datasets.}
Prior work has developed a range of datasets to support motion synthesis and understanding research.
Single-human motion--language datasets such as KIT~\cite{Plappert2016} and HumanML3D~\cite{guo2022cvpr_diverse} have been widely used for text-to-motion tasks, with InterGen~\cite{liang2024intergen}, Inter-X~\cite{xu2024interx}, and ReMoCap~\cite{ghosh2024remos} further extending this paradigm to human--human interaction.
Music-to-dance datasets such as AIST++~\cite{li2021ai}, AIOZ-GDANCE~\cite{le2023music}, and FineDance~\cite{li2023finedance} provide synchronized music and motion at scale, but depict either solo dancing or group choreography without close-contact partner interaction.
In duet settings, DD100~\cite{siyao2024duolando} and InterDance~\cite{li2024interdance} introduce duet dance datasets with paired motion and music. Across these resources, annotations are typically action labels or generic captions, and lack move-level labels and dense textual descriptions to bridge interactive motion and language.



In summary, generated dyadic interactions lack realism due to poor relative motion generation, as well as trade-offs between fine-grained details and long-sequence coherence. Multimodal language models show promise for addressing this task, yet current models require text as input, rather than solely motions tokens useful for nonverbal interactions. Indeed, human motion datasets that would be needed to train large language models lack annotation data for large-scale fine-tuning. To address these gaps, we propose SalsaAgent, a multimodal embodied LLM that aims to generate high-quality reactive motion as a kinematic reference for retargeting to socially interactive embodied agents. SalsaAgent uses motion tokens as both input and output to support nonverbal communication, enabled by (1) the addition of full-body and relative motion tokens, (2) the use of MotionScript\cite{yazdian2023motionscript}, an automatic method for annotating human motions for LLM training, and (3) a two-phase token-to-diffusion approach for improved generation of follower dance.




    \section{Method}

We study reactive leader-to-follower social dance generation.
Given an audio sequence $\mathbf{a}=\{a^i\}_{i=0}^{T_a-1}$ and an observed leader motion sequence $\mathbf{x}_L=\{x_L^i\}_{i=0}^{T-1}$, our goal is to generate the corresponding follower motion sequence $\mathbf{x}_F=\{x_F^i\}_{i=0}^{T-1}$ by modeling $p_{\theta}(\mathbf{x}_F \mid \mathbf{a}, \mathbf{x}_L)$.
To achieve this objective, our method consists of three components: (1) a full-body and pair-wise relation tokenizers that discretize motion and interaction trajectories into compact tokens~\cite{yu2025socialgen}; (2) a multimodal interactive motion--language model with an extended vocabulary that aligns motion, relation, audio, and text tokens~\cite{jiang2024motiongpt}, trained with instruction tuning; and (3) a diffusion-based refinement stage that improves geometric coordination and interaction realism in a shared two-person motion space~\cite{liang2024intergen}.
\begin{figure*}[t]
    \centering
    \includegraphics[width=\textwidth]{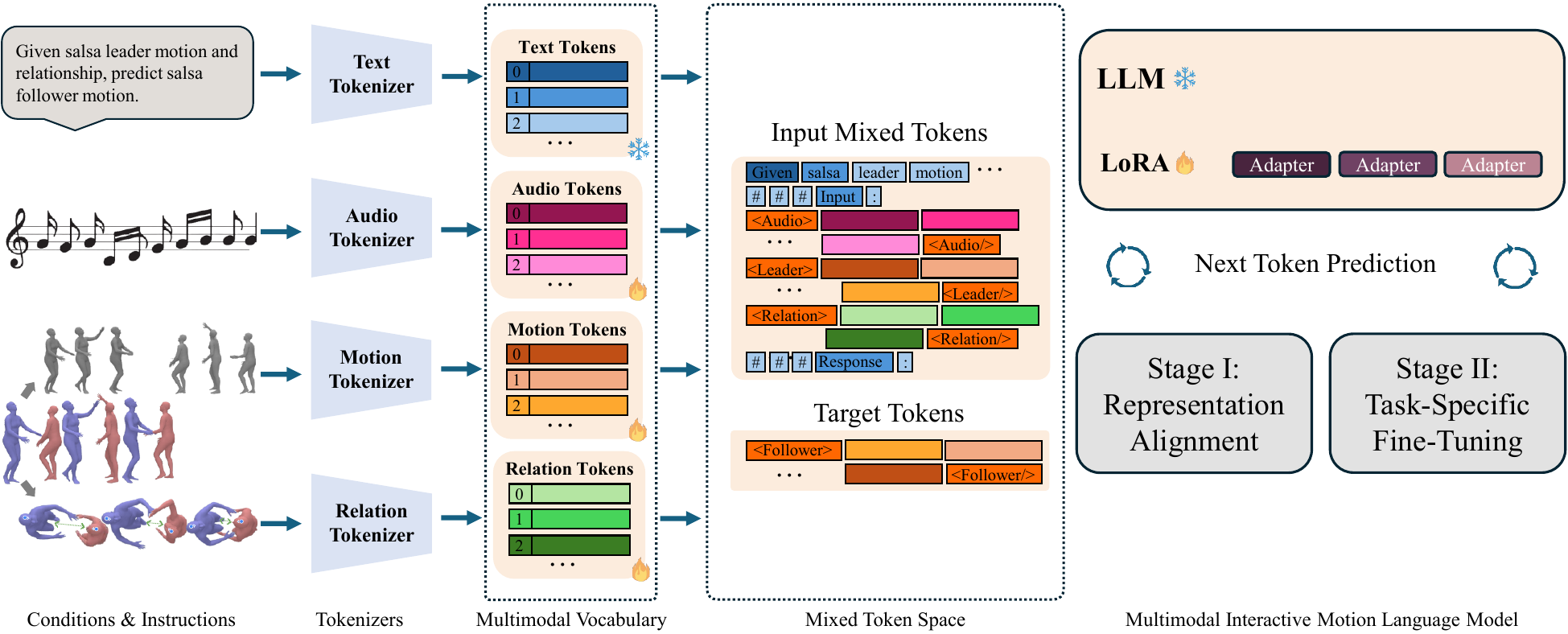}
    \caption{Overview of the SalsaAgent training pipeline: multimodal inputs (text, audio, leader/follower motion, and pair-wise relation) are tokenized and fused in a autoregressive LLM with an extended vocabulary; decoded motion passes through optional interaction-space diffusion refinement. During leader-to-follower inference, the model receives as input the instruction, initial relation tokens, a leader token sequence and audio token sequence. After token generation, diffusion-based refinement is applied.}
    \label{fig:salsa_agent_framework}
\end{figure*}

Figure~\ref{fig:salsa_agent_framework} summarizes the overall pipeline; the following details each component.

\textbf{Human interaction representation.}

We denote a two-person sequence by $\mathbf{x}=\{\mathbf{x}_L,\mathbf{x}_F\}$, where $L$ and $F$ indicate leader and follower, and $\mathbf{x}_h=\{x_h^i\}_{i=0}^{T-1}$ is the frame sequence for human $h\in\{L,F\}$.
This subsection defines per-frame motion states and pair-wise relation features used by the tokenizer in the next subsection.

\textit{Human motion representation.}
Root-centered canonical formats (e.g., HumanML3D~\cite{guo2022cvpr_diverse}) are effective for solo motion but lose explicit inter-person geometry.
For interaction, we adopt a non-canonical representation in the shared world-frame ~\cite{liang2024intergen}: for each human $h\in\{L,F\}$ and frame $i$,
\begin{equation}
x_h^i=\big[\,\mathbf{j}^{p,h}_{g},\,\mathbf{j}^{v,h}_{g},\,\mathbf{j}^{r,h},\,\mathbf{c}^{f,h}\,\big],
\end{equation}
where $\mathbf{j}^{p,h}_{g}\in\mathbb{R}^{3N_j}$ and $\mathbf{j}^{v,h}_{g}\in\mathbb{R}^{3N_j}$ are global joint \textit{positions} and \textit{velocities} in the world frame, $\mathbf{j}^{r,h}\in\mathbb{R}^{6N_j}$ are local joint \textit{rotations} in root space (6D representation), $\mathbf{c}^{f,h}\in\mathbb{R}^{4}$ denotes binary \textit{foot--ground contacts} from heel and toe velocities, and $N_j$ is the number of joints.

\textit{Pair-wise root relation.}
From this shared world-frame geometry, we define the relative placement of the two dancers over time: at frame $i$, the leader--follower root displacement on the ground plane and relative yaw are
\begin{equation}
\mathbf{r}^i=\left(r^x,r^z,r^\theta\right).
\end{equation}
where $(r^x,r^z)\in\mathbb{R}^2$ denote XZ-plane root offsets and $r^\theta\in[-\pi,\pi]$ is the relative yaw about the Y-axis.
We thus represent a two-person interaction with $\{x_L^i\}_{i=0}^{T-1}$, $\{x_F^i\}_{i=0}^{T-1}$, and $\{\mathbf{r}^i\}_{i=0}^{T-1}$, which are discretized into motion and relation tokens in the next subsection.

\textbf{Full-body and pair-wise relation tokenization.}
To interface with LLMs, we convert full-body motion and pair-wise relation trajectories into discrete tokens using vector-quantized variational autoencoders (VQ-VAEs)~\cite{van2017neural}. In contrast to frame-level quantization~\cite{jiang2024motiongpt, zhang2024motiongpt, zhou2023ude}, we tokenize short temporal windows to capture dynamic motion patterns~\cite{Gesture2Vec, yu2025socialgen}.

First, we split motion and relation trajectories into fixed-length windows of $\tau$ frames (we use $\tau=20$). For each leader/follower motion window $\mathbf{X}_h^{w}\in\mathbb{R}^{\tau\times D_x}$, we apply a rigid canonicalization transform such that the first frame is centered at the XZ origin and faces forward. 
The corresponding relation window $\mathbf{r}^{w}\in\mathbb{R}^{\tau\times D_r}$ preserves the relative spatial configuration and enables inverse rigid transforms to recover trajectories in the shared world-frame.

Next, each VQ-VAE consists of an encoder $\mathcal{E}$, a decoder $\mathcal{D}$, and a learnable codebook $\mathbf{C}=\{\mathbf{c}_k\}_{k=1}^{K}$~\cite{van2017neural, jiang2024motiongpt}. For an input window (either $\mathbf{X}_h^{w}$ or $\mathbf{r}^{w}$), we compute the encoded feature $\mathbf{z}$ and select the nearest codebook entry:
\begin{equation}
k^*=\arg\min_{k\in\{1,\ldots,K\}}\lVert \mathbf{z}-\mathbf{c}_k\rVert_2,\qquad \hat{\mathbf{z}}=\mathbf{c}_{k^*}.
\end{equation}
The index $k^*$ serves as the discrete token for the language model, while $\mathcal{D}$ reconstructs the input window from $\hat{\mathbf{z}}$. Following prior motion tokenization approaches~\cite{jiang2024motiongpt, zhang2023generating, zhang2024motiongpt}, both encoders and decoders are implemented with GRU-based recurrent networks~\cite{Gesture2Vec}. Each tokenizer is optimized with a combination of reconstruction, and commitment loss~\cite{zhang2023generating}; codebook embeddings are updated via exponential moving average EMA~\cite{razavi2019generating}, while inactive codes are reset~\cite{zhang2023generating}.

We instantiate two such VQ-VAEs for full-body motion and pair-wise relation windows, with separate codebooks $\mathbf{C}_m$ and $\mathbf{C}_r$ of sizes $K_m$ and $K_r$, resulting in token sequences $k^m$ and $k^r$, respectively. After pretraining, both tokenizers are frozen during LLM training.

\textbf{Multimodal Interactive Motion Language Model.}

\textit{Multimodal vocabulary and prompts.}
We adapt a pretrained autoregressive LLM to the multimodal setting by extending its vocabulary to integrate motion, relation, audio, and text tokens within a single sequence.
Motion and relation indices $k_h^m$ and $k^r$ are produced by the frozen VQ-VAEs, and audio tokens by a pretrained wave tokenizer~\cite{Wavtokenizer} to indices $k^a\in\{1,\ldots,K_a\}$.
We map codebook indices in correspondence with LLM token types, defining motion, relation, and audio vocabularies as $\mathcal{V}_m=\{\langle \mathrm{M}_i\rangle\}_{i=1}^{K_m}$, $\mathcal{V}_r=\{\langle \mathrm{R}_i\rangle\}_{i=1}^{K_r}$, and $\mathcal{V}_a=\{\langle \mathrm{A}_i\rangle\}_{i=1}^{K_a}$, where $\langle \mathrm{M}_i\rangle$ is equivalent to $k_h^m=i$, and similarly for relation and audio.
We also introduce special tokens including role markers (e.g., \texttt{<Follower>}, \texttt{</Follower>}) and paired tokens that explicitly indicate the start and end of each modality. Base text tokens use the original tokenizer unchanged.
A typical training input follows the layout: \texttt{<SYS> <Audio> <A\_*> </Audio> <Leader> <M\_*> </Leader> <Relation> <R\_*> </Relation> <Follower> <M\_*> </Follower>}.
The model thus conditions on audio, leader motion, and relation to predict output indices.

\textit{Training objective.}
We represent each training example as a single interleaved token sequence $y_{1:N}$ containing task instructions, multimodal conditioning tokens, and supervised target output tokens. We train the autoregressive LLM with next-token prediction $p_{\theta}(y_t \mid y_{<t})$, where all information required to predict $y_t$ is contained in prior tokens $y_{<t}$. Training minimizes the negative log-likelihood on the supervised target outputs:
\vspace{-15pt}
\begin{equation}
\mathcal{L}_{\mathrm{LLM}}=-\sum_{t=1}^{N}\log p_{\theta}(y_t\mid y_{<t}).
\end{equation}
\vspace{-10pt}
\\
where non-target positions are masked out in loss calculation.

\textit{LoRA Training.}
We use Low-Rank Adaptation (LoRA)~\cite{hu2021lora} for multimodal interactive motion--language learning, preserving general language priors while introducing only a small set of trainable parameters. During training, we fix the original text tokenizer, text-token embeddings, and output head, and train only the newly introduced multimodal token embeddings together with LoRA adapters in the LLM.

\textit{Two-stage fine-tuning.}
We use a two-stage training strategy for multimodal interactive motion--language learning.
In Stage~I (representation alignment), we optimize $\mathcal{L}_{\mathrm{LLM}}$ on a diverse set of supervised tasks including role-conditioned generation, motion completion, cross-role prediction, and text--motion tasks to align newly introduced multimodal tokens with the pretrained language space. 
For each task, prompts combine task instructions, metadata such as style captions and move labels when available, and fine-grained MotionScript~\cite{yazdian2023motionscript} captions for text--motion related tasks, followed by multimodal input and output tokens.
MotionScript captions emphasize fine-grained limb configuration and local body motion rather than abstract move-level natural language; they are used as training supervision for text--motion alignment and are not required as user input for leader--follower inference.
In this stage, both LoRA adapters and multimodal token embeddings are trainable. In Stage~II (task-specific fine-tuning), we freeze multimodal token embeddings and continue optimizing only LoRA adapters for the target motion--language task (i.e., leader-to-follower generation) with the same objective $\mathcal{L}_{\mathrm{LLM}}$.

\textit{Inference.}
At test time, we follow the same prompt layout as in Stage II training: observed audio, leader motion, an initial relation token, and a style caption. MotionScript is not used at inference. The model autoregressively predicts next relation tokens and follower motion tokens until the end token (e.g., \texttt{</Follower>}). 
For reconstruction we use only the initial relation and update partner geometry from the decoded follower to avoid mismatch between decoded relation and motion.


\textbf{Diffusion-Based Refinement Stage.}
Although LLM-generated and VQ-decoded outputs are semantically coherent, geometric artifacts can remain (e.g., partner-distance drift, contact mismatch, and local footstep artifacts). To improve interaction-level precision, we apply a diffusion-based refinement prior in shared world-frame motion~\cite{tevet2022human, liang2024intergen}.
We represent a two-person trajectory as
\begin{equation}
\mathbf{X}=[\mathbf{x}_L,\mathbf{x}_F].
\end{equation}
Let $\tilde{\mathbf{X}}^{(t)}$ denote the noisy state at diffusion step $t\in\{1,\ldots,N_{\mathrm{step}}\}$ with noise level $\sigma_t$. We train a denoiser network $D_{\theta}$ with the objective
\begin{equation}
\mathcal{L}_{\mathrm{diff}}
=\mathbb{E}_{\mathbf{X},t,\epsilon}
\left[\lambda_t\left\|\mathbf{X}-D_{\theta}\!\left(\mathbf{X}+\sigma_t\epsilon,\,t,\,\mathbf{c}\right)\right\|_2^2\right],
\end{equation}
where $\epsilon\sim\mathcal{N}(0,\mathbf{I})$, $\lambda_t$ is a timestep-dependent weight, and $\mathbf{c}$ is conditioning context (e.g., style metadata).


At inference, given observed leader motion $\mathbf{x}_L$ and LLM output $\hat{\mathbf{x}}_F$, we initialize
$\hat{\mathbf{X}}=[\mathbf{x}_L,\hat{\mathbf{x}}_F]$.
Then, we add noise only to the follower branch up to step $t_r$, then run reverse denoising from $t=t_r$ to $t=0$ while fixing the leader branch:
\begin{equation}
\tilde{\mathbf{X}}^{(t)}=[\mathbf{x}_L,\tilde{\mathbf{x}}_F^{(t)}].
\end{equation}
The refined follower motion is $\tilde{\mathbf{x}}_F^{(0)}$. In practice, we use a short refinement schedule (e.g., $t_r=10$ with $N_{\mathrm{step}}=50$), which improves geometric consistency while preserving the generated motion semantics.
    \section{Experiments}

We evaluate SalsaAgent on the CoMPAS3D~\cite{burkanova2025compas3d} dataset to answer three questions: (i) whether SalsaAgent improves objective duet-generation performance over strong baselines, (ii) how each proposed component contributes to those gains, and (iii) whether humans prefer SalsaAgent outputs in blind comparisons.

\textbf{Dataset}. We evaluate on CoMPAS3D~\cite{burkanova2025compas3d}, a motion-captured corpus of improvised partner salsa with 72 salsa duet recordings (about 2.5 minutes each), close leader--follower interaction, synchronized music, and high-fidelity 3D motion.
The dataset consists of over three hours from nine pairs (18 dancers) and three proficiency levels, and includes frame-level expert annotations of salsa move types, stylistic variation, and execution errors aligned with captured motion.
We annotate both dancers with rule-based MotionScript~\cite{yazdian2023motionscript} algorithm, creating about $300$K captions from $5$,s windows with a $2.5$,s stride ($50\%$ overlap). An example is: ``He moves a great distance backward rapidly, and turns clock-wise. His right hand moves to the left hand \dots''
We focus on \textbf{leader-to-follower} generation conditioned on observed leader motion and music, and report test-set metrics in aggregate and by proficiency.
All models are trained and evaluated under the same official train/validation/test split as in dataset release, with shared preprocessing and windowing as Sec.~III.

\textbf{Baselines}. We compare SalsaAgent against \textit{Duolando}\cite{siyao2024duolando}, \textit{InterGen}\cite{liang2024intergen}, and \textit{InterMask}~\cite{javed2024intermask}, and report Groundtruth as reference. We retrain all three baselines on CoMPAS3D under the same official split, with matched conditioning where applicable; as InterGen and InterMask do not accept music, follower motion is generated from leader motion only.


\textbf{Objective evaluation.} Following prior duet-dance evaluation~\cite{siyao2024duolando}, we assess four aspects.
For \emph{individual follower quality}, we use Fr\'echet Inception Distance (FID)~\cite{heusel2017gans} and diversity (Div) in two feature spaces, where subscripts $k$ and $g$ denote kinematic~\cite{onuma2008fmdistance} and graphical~\cite{muller2005efficient} features, respectively.
FID measures the distribution gap between generated and reference motions, and Div measures variability across generated samples.
For \emph{interaction quality}, we follow~\cite{siyao2024duolando} and compute cross-distance features between leader and follower joints, then evaluate them with FID$_{cd}$ and Div$_{cd}$.
For \emph{rhythmic synchrony}, we report BED for leader--follower beat synchrony~\cite{siyao2024duolando} and BAS for motion--music alignment~\cite{siyao2022bailando}.
For BAS, closer to the ground-truth value ($\rightarrow$) is preferred, due to salsa syncopated (off-beat) rhythms\cite{fitch2016dance}.
For \emph{contact and physical plausibility}, we report kinematic proxies for foot-slip, handhold proximity, and collision\cite{yang2025omniretarget}. Specifically, we measure foot skating rate (FSR)\cite{siyao2024duolando}, the fraction of frames where the body translates horizontally while a foot remains nearly static relative to the pelvis; Hand-CF, the fraction of frames with minimum leader--follower wrist distance $<\tau{=}15$,cm; and PR, the fraction of frames with inter-person mesh penetration. For PR, closer to the ground-truth value ($\rightarrow$) is preferred, since minimizing penetration alone can favor pairs that remain farther apart (lower Hand-CF).
Metric directionality is shown in Table \ref{tab:main_quant}, where higher ($\uparrow$), lower ($\downarrow$), or closer to the ground-truth ($\rightarrow$) is better.

\textbf{Subjective evaluation.} Following university ethics board approval, we recruited participants from Prolific for the subjective evaluation. Participants first reviewed a consent form, confirmed at least one year of prior dance experience in any type of dance, and completed a headphone check. In each trial, participants evaluated four methods side-by-side (Groundtruth, Duolando, InterGen, and SalsaAgent) presented in anonymized and randomized order. They rated each method on a 1--5 Likert scale (1: very low, 5: very high) for each of six competitive salsa evaluation dimensions comprising timing, musicality, technique, difficulty, partner coordination, and originality \cite{canadasalsacongress_rules_2024}. Participants then ranked the four methods from best (1st) to worst (4th). Definitions for each rating dimension were provided during evaluation and reflect dimensions used in salsa competition \cite{canadasalsacongress_rules_2024}. 


\textbf{Implementation details.} We tokenize with 20-frame segments (1 s at 20 fps) and train separate VQ-VAEs~\cite{van2017neural,zhang2023generating} for motion and relation, then freeze both for LLM tuning.
Similar to Gesture2Vec~\cite{Gesture2Vec}, we use two-layer bidirectional GRU encoder and directional GRU decoder networks for sequence-level encoding and reconstruction.
The motion tokenizer uses latent size 512 and 512 codes, trained for 2000 epochs (batch 2048, lr $10^{-4}$); the relation tokenizer uses latent size 32 with the 512 codebook size, same batch size and learning rate, trained for 200 epochs.
Both use EMA-reset quantization ($\mu=0.95$), commitment weight 0.02, velocity loss weight 0.1, 5 warmup epochs, and scheduler gamma 0.05.
We adopt Gemma2-2B-it~\cite{gemma2024report} as the LLM backbone and apply LoRA for parameter-efficient tuning~\cite{hu2021lora} (rank 64, alpha 64, dropout 0.1): Stage I trains multimodal embeddings and LoRA on multi-task data with learning rate $2\times10^{-5}$, and Stage II freezes added embeddings and fine-tunes only LoRA for 100 epochs (batch size 4) with learning rate $1\times10^{-5}$.
We train a conditional diffusion-based denoiser for motion refinement using a frozen CLIP ViT-L/14 text encoder~\cite{radford2021learning}, with style provided as a text caption for conditioning. The model is optimized with 1000 diffusion training steps and employs DDIM sampling~\cite{song2020denoising} with 50 inference steps ($\eta = 0$). A cosine noise schedule is adopted~\cite{nichol2021improved}, and classifier-free guidance is applied with a 10\% conditioning dropout rate and a guidance scale of 3.5.
All models are trained using the AdamW optimizer~\cite{loshchilov2017decoupled}, and experiments are run on a single NVIDIA RTX A6000 GPU.
    \section{Results and Discussion}


Table~\ref{tab:main_quant} reports Sec.~IV-C metrics on CoMPAS3D for ground truth, InterGen, Duolando, InterMask, SalsaAgent ablations, and full SalsaAgent.

\newcommand{\et}[2]{$#1_{{\scriptstyle\pm#2}}$}
\begin{table*}[t]
  \centering
  \caption{Quantitative comparison on CoMPAS3D. We include ground truth as reference, baselines, ablations, and full SalsaAgent. We present solo, interactive, motion--music alignment, and contact metrics. $\pm$ denotes a 95\% bootstrap confidence interval over 100 runs. FSR is reported in percent. Arrows indicate whether higher ($\uparrow$), lower ($\downarrow$), or closer to the ground-truth ($\rightarrow$) is better. Among \emph{generative} methods (excluding ground truth), the best and second-best value in each column is shown in \textbf{bold} and \underline{underlined}, respectively.}
  \label{tab:main_quant}
  \resizebox{\textwidth}{!}{%
  \setlength{\tabcolsep}{0.9mm}%
  \scriptsize
  \begin{tabular}{l ccccc ccc c cc}
    \toprule
     & \multicolumn{5}{c}{Solo Metrics}
     & \multicolumn{3}{c}{Interactive Metrics}
     & \multicolumn{1}{c}{Rhythmic}
     & \multicolumn{2}{c}{Contact Proxies} \\
    \cmidrule(r){2-6} \cmidrule(lr){7-9} \cmidrule(lr){10-10} \cmidrule(l){11-12}
    Method
      & FID$_k$($\downarrow$) & FID$_g$($\downarrow$) & Div$_k$($\uparrow$) & Div$_g$($\uparrow$) & FSR(\%)($\downarrow$)
      & FID$_{cd}$($\downarrow$) & Div$_{cd}$($\uparrow$) & BED($\uparrow$)
      & BAS($\rightarrow$)
      & Hand-CF($\uparrow$) & PR($\rightarrow$) \\
    \midrule
    Groundtruth
      & \et{0.00}{.00} & \et{0.00}{.00} & $8.741$ & $0.809$ & $0.14$
      & \et{0.00}{.00} & $2.0111$ & $0.4206$
      & $0.1552$
      & $0.2645$ & $0.5162$ \\
    \midrule
    InterGen
      & \et{24.58}{.75} & \et{13.57}{.45} & \bm{$8.594$} & $0.720$ & $3.31$
      & \et{18.62}{.28} & \bm{$2.07$} & $0.3163$
      & \underline{$0.1970$}
      & $0.1302$ & $0.6480$ \\
    Duolando
      & \et{12.46}{.45} & \et{15.54}{.25} & \underline{$7.964$} & \bm{$0.782$} & $29.68$
      & \et{20.14}{.23} & \underline{$1.93$} & $0.3045$
      & \bm{$0.2201$}
      & \bm{$0.2130$} & $0.7798$ \\
    InterMask
      & \et{11.55}{.15} & \et{12.85}{.15} & $8.20$ & $0.745$ & $6.47$
      & \et{13.90}{.28} & $1.98$ & $0.3120$
      & $0.2010$
      & $0.0404$ & \bm{$0.4804$} \\
    \midrule
    SalsaAgent w/o LLM (GPT-2)
      & \et{10.37}{.39} & \et{14.37}{.49} & $6.90$ & $0.740$ & $3.83$
      & \et{9.20}{.20} & $1.85$ & $0.3300$
      & $0.1980$
      & $0.0300$ & $0.7100$ \\
    SalsaAgent w/o Audio
      & \et{5.81}{.12} & \et{13.21}{.14} & $6.089$ & \underline{$0.761$} & $2.93$
      & \underline{\et{5.73}{.14}} & $1.81$ & \bm{$0.3605$}
      & $0.1840$
      & $0.1251$ & $0.3800$ \\
    SalsaAgent w/o MotionScript
      & \et{7.26}{.28} & \et{16.87}{.33} & $6.318$ & $0.734$ & \underline{$2.61$}
      & \et{5.78}{.13} & $1.88$ & \underline{$0.3592$}
      & $0.1942$
      & $0.1500$ & $0.5790$ \\
    SalsaAgent w/o Relation
      & \et{6.53}{.10} & \et{14.32}{.15} & $6.562$ & $0.716$ & $2.94$
      & \et{7.76}{.15} & $1.91$ & $0.3528$
      & $0.1855$
      & $0.0900$ & $0.4640$ \\
    SalsaAgent w/o Refinement
      & \bm{\et{4.84}{.09}} & \bm{\et{11.92}{.19}} & $4.951$ & $0.692$ & $4.62$
      & \et{8.32}{.18} & $1.78$ & $0.2386$
      & $0.1765$
      & $0.0800$ & $0.4400$ \\
    \midrule
    SalsaAgent (ours)
      & \underline{\et{5.21}{.07}} & \underline{\et{12.48}{.21}} & $6.241$ & $0.728$ & \bm{$2.16$}
      & \bm{\et{5.26}{.13}} & $1.84$ & $0.3567$
      & $0.1904$
      & \underline{$0.1824$} & \underline{$0.5610$} \\
    \bottomrule
  \end{tabular}%
  }
\end{table*}

Relative to  InterGen~\cite{liang2024intergen}, Duolando~\cite{siyao2024duolando}, and InterMask\cite{javed2024intermask}, SalsaAgent is strongest on partner coordination and follower kinematics, while not maximizing every diversity, music-alignment, or contact proxy (Table \ref{tab:main_quant}).

In follower quality, SalsaAgent achieves $FID_k{=}5.21$ and $FID_g{=}12.48$.
Compared with Duolando, this is a $58\%$ reduction in $FID_k$ and a lower $FID_g$; compared with InterGen, $FID_k$ drops by $79\%$ and $FID_g$ is lower; compared with InterMask, $FID_k$ drops by $55\%$, and $FID_g$ is lower.
Since $FID_k$ and $FID_g$ are computed from kinematic and trajectory-based motion features~\cite{onuma2008fmdistance,muller2005efficient}, these reductions indicate closer matching to reference follower dynamics and global motion patterns.
SalsaAgent does not lead $Div_k$/$Div_g$, likely because unified whole-body tokenization concentrates generation more than Duolando's part-wise codebook composition and InterGen's raw-space diffusion, and as noted in prior work, higher diversity primarily reflects spread rather than guaranteed coverage or quality~\cite{naeem2020reliable}.

For interaction quality, SalsaAgent has the lowest $FID_{cd}$ ($5.26$), versus $18.62$ for InterGen, $20.14$ for Duolando, and $13.90$ for InterMask.
Because $cd$ features summarize leader--follower pairwise geometry~\cite{siyao2024duolando}, this indicates closer matching of partner-relative spacing to the reference distribution.
InterGen records the highest $Div_{cd}$; we interpret this as somewhat wider variation in cross-distance patterns across samples; however, diversity appears to come at the expense of FID.
SalsaAgent also improves $BED$, suggesting stronger leader--follower motion alignment and beat synchrony.

For motion--music alignment, 
SalsaAgent ($0.190$) is closer to groundtruth BAS ($0.155$) while InterGen ($0.197$), InterMask ($0.201$), and Duolando ($0.220$) obtain higher $BAS$ values.
Moreover, SalsaAgent achieves the lowest FSR ($2.16\%$), well below Duolando ($29.68\%$), InterGen ($3.31\%$), and InterMask ($6.47\%$). Duolando leads Hand-CF among generative methods ($0.213$), with SalsaAgent second ($0.182$); for PR ($\rightarrow$), InterMask is closest to ground truth ($0.480$ vs.\ $0.516$), then SalsaAgent ($0.561$), reflecting the proximity--penetration trade-off in Sec.IV. 
Together, these results indicate that SalsaAgent's main gains concentrate on partner geometry and kinematic fidelity, with a $BAS$ closer to ground-truth.

\textbf{Qualitative Results.}\begin{figure*}[t]
    \centering
    \includegraphics[width=\textwidth]{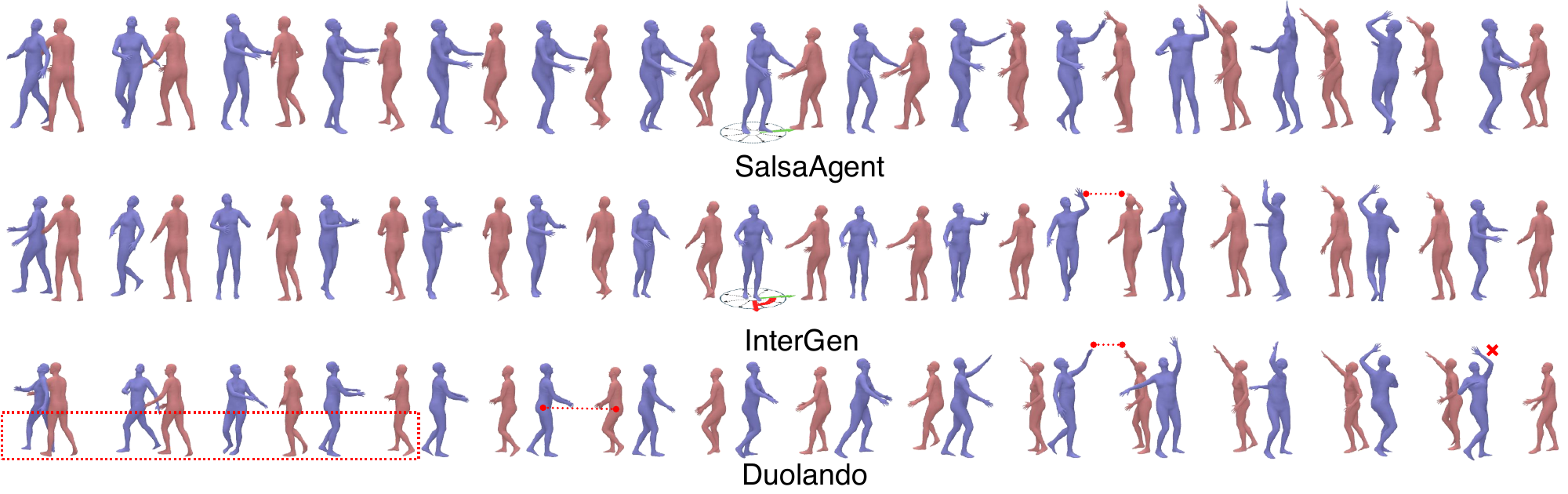}
    \caption{Qualitative comparison against baseline methods. Given the same leader motion (red), we compare generated follower motion (blue) from SalsaAgent, Duolando, and InterGen. SalsaAgent maintains correct orientation (top), while InterGen and Duolando lose hand contact (middle, bottom). Duolando also suffers from foot-skating artifacts (bottom left).}
    \label{fig:qual_compare}
\end{figure*}

Figure~\ref{fig:qual_compare} shows qualitative comparisons of generated follower motion conditioned on a given leader, across SalsaAgent and baselines.
In SalsaAgent, the follower maintains more consistent relative orientation to the leader and a more stable dance ``frame,'' aligned with stronger interaction metrics in Table~\ref{tab:main_quant} (notably $FID_{cd}$ and $BED$).
Compared to Duolando, we observe fewer foot-skating artifacts, likely due to stitching body-part tokens with independent root translation~\cite{siyao2024duolando}, while SalsaAgent uses a single whole-body tokenization approach. Compared to InterMask, which uses whole-body tokenization, and InterGen a diffusion model over raw motion~\cite{liang2024intergen}, we observe a looser duet frame and less coordinated interactions, suggesting the importance of explicit relation tokens in our pipeline.

Failure cases occur under abrupt changes in orientation and distance, where relative-orientation drifts may appear. Refinement stage reduces this effect but does not fully remove it. Aggregated error from imperfect motion VQ reconstruction may occasionally lead to misalignment between decoded trajectories and relation-token guidance.

\textbf{Human study results}. We further obtain subjective ratings of samples generated using each method from 31 participants aged 18--51 years (mean $30.4$; 18 male, 13 female). Participants compared methods across 12 trials of four-way stimulus comparisons, providing 372 evaluations per method. Stimuli for each method were presented as a balanced Latin Square to eliminate ordering effects. 
Figure~\ref{fig:subjective_chart} summarizes subjective results for each of six Likert criteria defined in Sec.~IV across Groundtruth (GT), SalsaAgent (SA), Duolando (SD), and InterGen (SI); bars show mean$\pm$SD for readability. Significance brackets are computed using pairwise Wilcoxon signed-rank tests. 
As expected, GT achieved the highest scores, with all six Likert criteria revealing statistically significant differences between GT and each generated method ($p<0.05$). 

In terms of ranking, we observe a significant preference for SalsaAgent. SA achieved the best average rank among trainable methods (2.505 vs. 2.866/2.892, lower is better) and was selected as top-1 in 21.5\% of trials, compared with 13.7\% for Duolando and 10.8\% for InterGen.
We completed a Chi-square goodness-of-fit test on top-choice counts and found a significant preference difference among methods ($\chi^2(3, N=372)=176.41$, $p<0.001$), consistent with participants favoring SalsaAgent over Duolando and InterGen.
These results align with trends in Table~\ref{tab:main_quant}, where SalsaAgent is rated higher than other baselines across factors.

\begin{figure}[t]
\centering
\includegraphics[width=0.9\columnwidth]{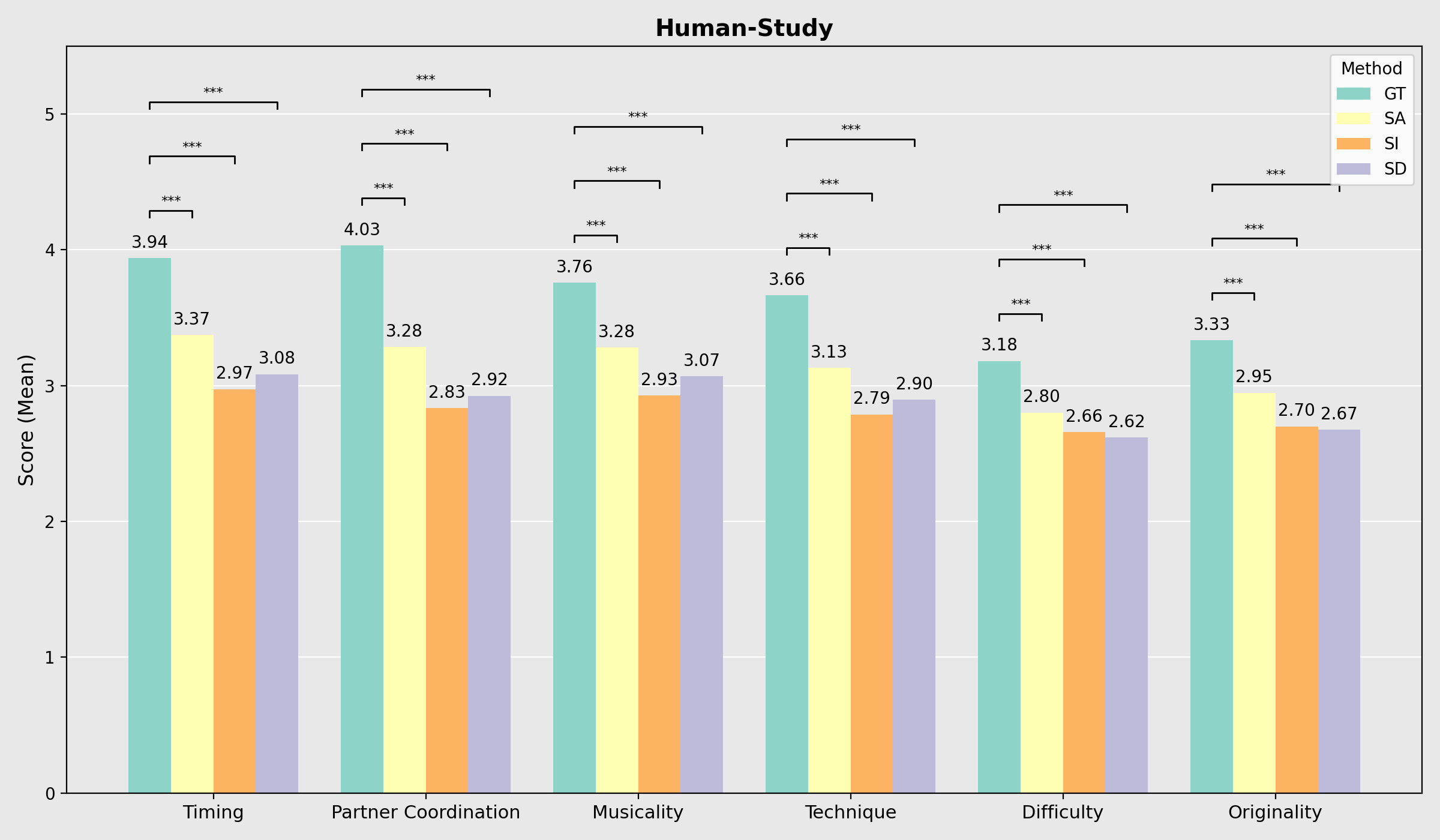}
\caption{User study results. Each column shows descriptive mean on a 1--5 scale for Groundtruth (GT), SalsaAgent (SA), Duolando (SD), and InterGen (SI). Significance brackets show pairwise Wilcoxon signed-rank tests against GT (* $p<0.05$, ** $p<0.01$, *** $p<0.001$). Overall, SalsaAgent received the highest subjective ratings among trainable methods.}
\label{fig:subjective_chart}
\end{figure}

\textbf{Ablation studies.} 
\label{subsec:ablation}
We ablate each major componentin Table~\ref{tab:main_quant} (above the full SalsaAgent row), replacing the LLM with a transformer architecture (GPT-2\cite{radford2019language}) trained from the scratch at a similar number of trainable parameters, while the others remove one conditioning path or the diffusion refinement stage. For all variants, we keep all other training settings matched (Sec.~IV).

\textit{Removing language prior.}
Replacing Gemma with GPT-2 under the same multimodal vocabulary strongly degraded follower quality, partner geometry, and contact proxies.
This suggests that the knowledge embedded through LLM pretraining contributes to nonverbal token prediction. 

\textit{Removing audio.}
Interestingly, \emph{SalsaAgent w/o audio} recorded the highest $BED$ in the grid. This can be explained by the follower better aligning to leader timing when not coupled to strict music--beat tracking.

\textit{Removing MotionScript.}
Removing MotionScript yielded the largest $FID_g$ spike, indicating strongly coupled language grounding to motion is an important factor for training.

\textit{Removing relation features.}
Without relation tokens, interaction metrics degraded, supporting explicit relation conditioning for partner geometry~\cite{siyao2024duolando}.
$FID_k$ and $FID_g$ declined, showing follower quality depends on interaction context.

\textit{Removing refinement.}
Without diffusion refinement, interaction quality degraded most strongly,
while solo follower FIDs slightly improved, showing a trade-off that still favors refinement for partner coordination and foot skating.
Overall, the full model remains most balanced across these metrics.

\textbf{Inference Latency.} We report inference latency for 100-frame sequences, averaged over 100 inference runs on an RTX A6000.
Duolando, InterMask, InterGen, and SalsaAgent require on average $0.32$\,s, $0.92$\,s, $1.13$\,s, and $1.08$\,s, respectively.
These delays suit upstream kinematic motion reference generation~\cite{yang2025omniretarget}, but remain too high for real-time interactive leader--follower control, motivating future work on accelerating autoregressive LLM decoding~\cite{cao2025real} and faster alternatives to iterative diffusion refinement~\cite{lee2026implicit}.

\subsection{Limitations and Future Work}

Subjective and objective evidence together support SalsaAgent as a strong nonverbal follower motion generator on CoMPAS3D.
Several limitations remain, including that no generative configuration achieves the best score in every metric. Our contact measures are kinematic proxies and do not reflect balance or underlying physical dynamics. Results are specific to CoMPAS3D salsa duets as a challenging testbed for multimodal human--human interaction and require further evaluation in other interaction settings. Upon close inspection of generated samples, we also noted that whole-body motion tokenization can lose fine-grained limb and footwork detail, and that longer generated sequences more often reveal aggregated misalignment between motion-token reconstruction and relation-token guidance.

Future work includes conditioning the diffusion prior on audio to capture high-frequency timing cues and to narrow the $BAS$ gap, hierarchical or body-part VQ tokenization to reduce whole-body quantization loss to preserve fine-grained motion detail, and joint optimization of motion and relation VQ-VAEs under cross-modal consistency losses to improve alignment between decoded trajectories and predicted interaction geometry.
Additional directions include physics-aware decoding and articulated mesh representations to improve contact and weight transfer during fast movements.
    \section{Conclusion}

In this work, we study music-conditioned leader-to-follower dance generation that can provide a kinematic motion reference toward socially interactive embodied agents.
We introduce SalsaAgent, a multimodal embodied language model for interactive dance generation that formulates partner interaction as nonverbal motion token passing and improves upon baseline methods in a salsa dance dataset. We use separate vector-quantized tokenizers for single-person motion and interaction relations, alongside audio tokens, and ground the motion tokens in natural language. Training follows a two-stage strategy: multimodal representation alignment followed by task-specific fine-tuning for leader-to-follower generation, followed by a lightweight diffusion-based refinement stage that improves fine-grained details such as contact and orientation.

Quantitative evaluation shows that our method achieves consistent improvements in objective metrics of follower realism and partner coordination, reflecting stronger kinematic plausibility and interaction coherence. Subjective evaluations further indicate a clear preference for our generated motions over competing approaches, while a gap remains compared to real reference performances.

    
\fi

\bibliographystyle{unsrt2authabbrvpp}
\bibliography{references}

@article{McMains02012016,
  author = {Juliet McMains},
  title = {Salsa Steps Toward Intercultural Education},
  journal = {J. Dance Educ.},
  volume = {},
  number = {},
  pages = {},
  year = {2016},
  publisher = {Routledge},
  doi = {10.1080/15290824.2015.1048865}
}

@book{hanna1987dance,
  title={To dance is human: A theory of nonverbal communication},
  author={Hanna, Judith Lynne},
  year={1987 },
  publisher={University of Chicago Press}
}

@inproceedings{xu2024interx,
  title={Inter-x: Towards versatile human-human interaction analysis},
  author={Xu, Liang and Lv, Xintao and Yan, Yichao and Jin, Xin and Wu, Shuwen and Xu, Congsheng and Liu, Yifan and Zhou, Yizhou and Rao, Fengyun and Sheng, Xingdong and others},
  booktitle={CVPR},
  pages={},
  year={2024}
}

@article{10.1145/3323335,
  author = {Raheb, Katerina El and Stergiou, Marina and Katifori, Akrivi and Ioannidis, Yannis},
  title = {Dance Interactive Learning Systems: A Study on Interaction Workflow and Teaching Approaches},
  year = {2019},
  publisher = {Association for Computing Machinery},
  address = {New York, NY, USA},
  journal = {ACM Comput. Surv.},
  volume = {},
  number = {},
  articleno = {},
  numpages = {},
  pages = {}
}

@inproceedings{siyao2024duolando,
  title={Duolando: Follower gpt with off-policy reinforcement learning for dance accompaniment},
  author={Siyao, Li and Gu, Tianpei and Yang, Zhitao and Lin, Zhengyu and Liu, Ziwei and Ding, Henghui and Yang, Lei and Loy, Chen Change},
  booktitle={ICLR},
  pages={},
  year={2024}
}

@inproceedings{li2024interdance,
  title={InterDance: Reactive 3D Dance Generation with Realistic Duet Interactions},
  author={Li, Ronghui and Zhang, Youliang and Zhang, Yachao and Zhang, Yuxiang and Su, Mingyang and Guo, Jie and Liu, Ziwei and Liu, Yebin and Li, Xiu},
  booktitle={{arXiv}},
  pages={},
  year={2025}
}

@inproceedings{yazdian2023motionscript,
  title={{MotionScript}: Natural language descriptions for expressive {3D} human motions},
  author={Yazdian, Payam Jome and Lagasse, Rachel and Mohammadi, Hamid and Liu, Eric and Cheng, Li and Lim, Angelica},
  booktitle={IROS},
  pages={},
  year={2025},
  organization={IEEE}
}

@inproceedings{siyao2022bailando,
  title={Bailando: 3d dance generation by actor-critic gpt with choreographic memory},
  author={Siyao, Li and Yu, Weijiang and Gu, Tianpei and Lin, Chunze and Wang, Quan and Qian, Chen and Loy, Chen Change and Liu, Ziwei},
  booktitle={CVPR},
  pages={},
  year={2022}
}

@article{le2023controllable,
  title={Controllable group choreography using contrastive diffusion},
  author={Le, Nhat and Do, Tuong and Do, Khoa and Nguyen, Hien and Tjiputra, Erman and Tran, Quang D and Nguyen, Anh},
  journal={TOG},
  volume={},
  number={},
  pages={},
  year={2023},
  publisher={ACM New York, NY, USA}
}

@inproceedings{le2023music,
  title={Music-driven group choreography},
  author={Le, Nhat and Pham, Thang and Do, Tuong and Tjiputra, Erman and Tran, Quang D and Nguyen, Anh},
  booktitle={CVPR},
  pages={},
  year={2023}
}

@inproceedings{guo2022cvpr_diverse,
  author    = {Guo, Chuan and Zou, Shihao and Zuo, Xinxin and Wang, Sen and Ji, Wei and Li, Xingyu and Cheng, Li},
  title     = {Generating Diverse and Natural {3D} Human Motions From Text},
  booktitle = {CVPR},
  month     = {June},
  year      = {2022},
  pages     = {}
}

@inproceedings{zhang2023generating,
  title={{T2M-GPT}: Generating Human Motion from Textual Descriptions with Discrete Representations},
  author={Zhang, Jianrong and Zhang, Yangsong and Cun, Xiaodong and Huang, Shaoli and Zhang, Yong and Zhao, Hongwei and Lu, Hongtao and Shen, Xi},
  booktitle={CVPR},
  year={2023}
}

@inproceedings{tevet2022human,
  title={Human Motion Diffusion Model},
  author={Tevet, Guy and Raab, Sigal and Gordon, Brian and Shafir, Yonatan and Bermano, Amit H and Cohen-Or, Daniel},
  booktitle={ICLR},
  year={2023}
}

@inproceedings{dabral2022mofusion,
  title={{MoFusion}: A Framework for Denoising-Diffusion-based Motion Synthesis},
  author={Dabral, Rishabh and Mughal, Muhammad Hamza and Golyanik, Vladislav and Theobalt, Christian},
  booktitle={CVPR},
  year={2023}
}

@inproceedings{tseng2022edge,
  title={{EDGE}: Editable Dance Generation From Music},
  author={Tseng, Jonathan and Castellon, Rodrigo and Liu, C Karen},
  booktitle={CVPR},
  pages={},
  year={2023}
}

@inproceedings{zhou2023ude,
  title={{UDE}: A Unified Driving Engine for Human Motion Generation},
  author={Zhou, Zixiang and Wang, Baoyuan},
  booktitle={CVPR},
  year={2023},
  pages={}
}

@inproceedings{guo2023momask,
  title={{MoMask}: Generative Masked Modeling of {3D} Human Motions},
  author={Guo, Chuan and Mu, Yuxuan and Javed, Muhammad Gohar and Wang, Sen and Cheng, Li},
  booktitle={CVPR},
  pages={},
  year={2024}
}

@article{van2017neural,
  title={Neural Discrete Representation Learning},
  author={van den Oord, Aaron and Vinyals, Oriol and Kavukcuoglu, Koray},
  journal={NeurIPS},
  volume={},
  number={},
  pages={},
  year={2017}
}

@inproceedings{razavi2019generating,
  title={Generating Diverse High-Fidelity Images with {VQ-VAE}-2},
  author={Razavi, Ali and van den Oord, Aaron and Vinyals, Oriol},
  booktitle={NeurIPS},
  volume={},
  pages={},
  year={2019}
}

@inproceedings{Gesture2Vec,
  title={Gesture2Vec: Clustering gestures using representation learning methods for co-speech gesture generation},
  author={Yazdian, Payam Jome and Chen, Mo and Lim, Angelica},
  booktitle={IROS},
  year={2022},
  organization={IEEE}
}

@inproceedings{hu2021lora,
  title={Lora: Low-rank adaptation of large language models},
  author={Hu, Edward J and Shen, Yelong and Wallis, Phillip and Allen-Zhu, Zeyuan and Li, Yuanzhi and Wang, Shean and Wang, Lu and Chen, Weizhu},
  booktitle={ICLR},
  pages={},
  year={2022}
}

@inproceedings{Wavtokenizer,
  title={Wavtokenizer: an efficient acoustic discrete codec tokenizer for audio language modeling},
  author={Ji, Shengpeng and Jiang, Ziyue and Wang, Wen and Chen, Yifu and Fang, Minghui and Zuo, Jialong and Yang, Qian and Cheng, Xize and Wang, Zehan and Li, Ruiqi and others},
  booktitle={ICLR},
  pages={},
  year={2025}
}

@article{jiang2024motiongpt,
  title={{MotionGPT}: Human Motion as a Foreign Language},
  author={Jiang, Biao and Chen, Xin and Liu, Wen and Yu, Jingyi and Yu, Gang and Chen, Tao},
  journal={NeurIPS},
  volume={},
  number={},
  pages={},
  year={2023}
}

@inproceedings{zhang2024motiongpt,
  title={{MotionGPT}: Finetuned {LLM}s are General-Purpose Motion Generators},
  author={Zhang, Yaqi and Huang, Di and Liu, Bin and Tang, Shixiang and Lu, Yan and Chen, Lu and Bai, Lei and Chu, Qi and Yu, Nenghai and Ouyang, Wanli},
  booktitle={AAAI},
  pages={},
  year={2024}
}

@article{brown2020language,
  title={Language Models are Few-Shot Learners},
  author={Brown, Tom and Mann, Benjamin and Ryder, Nick and Subbiah, Melanie and Kaplan, Jared and Dhariwal, Prafulla and Neelakantan, Arvind and Shyam, Pranav and Sastry, Girish and Askell, Amanda and others},
  journal={NeurIPS},
  volume={},
  number={},
  pages={},
  year={2020}
}

@article{touvron2023llama,
  title={{LLaMA}: Open and Efficient Foundation Language Models},
  author={Touvron, Hugo and Lavril, Thibaut and Izacard, Gautier and Martinet, Xavier and Lachaux, Marie-Anne and Lacroix, Timoth{\'e}e and Rozi{\`e}re, Baptiste and Goyal, Naman and Hambro, Eric and Azhar, Faisal and others},
  journal={arXiv},
  year={2023}
}

@inproceedings{wu2023nextgpt,
  title={{NExT-GPT}: Any-to-Any Multimodal {LLM}},
  author={Wu, Shengqiong and Fei, Hao and Qu, Leigang and Ji, Wei and Chua, Tat-Seng},
  booktitle={ICML},
  pages={},
  year={2024}
}

@InProceedings{han2023onellm,
  title={{OneLLM}: One Framework to Align All Modalities with Language},
  author={Han, Jiaming and Gong, Kaixiong and Zhang, Yiyuan and Wang, Jiaqi and Zhang, Kaipeng and Lin, Dahua and Qiao, Yu and Gao, Peng and Yue, Xiangyu},
  booktitle={CVPR},
  year={2024}
}

@inproceedings{liu2023interactive,
  title={Interactive Humanoid: Online Full-Body Motion Reaction Synthesis with Social Affordance Canonicalization and Forecasting},
  author={Liu, Yunze and Chen, Changxi and Yi, Li},
  booktitle={3DV},
  pages={},
  year={2025}
}

@inproceedings{ghosh2024remos,
  title={{ReMoS}: {3D} Motion-Conditioned Reaction Synthesis for Two-Person Interactions},
  author={Ghosh, Anindita and Dabral, Rishabh and Golyanik, Vladislav and Theobalt, Christian and Slusallek, Philipp},
  booktitle={ECCV},
  pages={},
  year={2024},
  organization={Springer}
}

@inproceedings{xu2024regennet,
  title={{ReGenNet}: Towards Human Action-Reaction Synthesis},
  author={Xu, Liang and Zhou, Yizhou and Yan, Yichao and Jin, Xin and Zhu, Wenhan and Rao, Fengyun and Yang, Xiaokang and Zeng, Wenjun},
  booktitle={CVPR},
  pages={},
  year={2024}
}

@article{sui2026survey,
  title={A survey on human interaction motion generation},
  author={Sui, Kewei and Ghosh, Anindita and Hwang, Inwoo and Zhou, Bing and Wang, Jian and Guo, Chuan},
  journal={IJCV},
  volume={},
  number={},
  pages={},
  year={2026},
  publisher={Springer}
}

@article{liang2024intergen,
  title={{InterGen}: Diffusion-Based Multi-Human Motion Generation Under Complex Interactions},
  author={Liang, Han and Zhang, Wenqian and Li, Wenxuan and Yu, Jingyi and Xu, Lan},
  journal={IJCV},
  volume={},
  number={},
  pages={},
  year={2024},
  publisher={Springer}
}

@inproceedings{javed2024intermask,
  title={{InterMask}: {3D} Human Interaction Generation via Collaborative Masked Modelling},
  author={Javed, Muhammad Gohar and Guo, Chuan and Cheng, Li and Li, Xingyu},
  booktitle={ICLR},
  pages={},
  year={2025}
}

@article{Plappert2016,
  author = {Matthias Plappert and Christian Mandery and Tamim Asfour},
  title = {The {KIT} Motion-Language Dataset},
  journal = {Big Data},
  publisher = {Mary Ann Liebert Inc},
  year = {2016},
  month = {},
  volume = {},
  number = {},
  pages = {},
  doi = {10.1089/big.2016.0028}
}

@inproceedings{yu2025socialgen,
  title={{SocialGen}: Modeling Multi-Human Social Interaction with Language Models},
  author={Yu, Heng and Zhang, Juze and Chen, Changan and Xiang, Tiange and Fang, Yusu and Niebles, Juan Carlos and Adeli, Ehsan},
  booktitle={3DV },
  year={2026}
}

@article{burkanova2025compas3d,
  title={{CoMPAS3D}: A Dataset and Benchmark for Interactive Motion},
  author={Burkanova, Bermet and Yazdian, Payam Jome and Zhang, Chuxuan and Evans, Trinity and Tutt{\"o}s{\'i}, Paige and Lim, Angelica},
  year={2025},
  journal={arXiv}
}

@inproceedings{girdhar2023imagebind,
  title={{ImageBind}: One Embedding Space to Bind Them All},
  author={Girdhar, Rohit and El-Nouby, Alaaeldin and Liu, Zhuang and Singh, Mannat and Alwala, Kalyan Vasudev and Joulin, Armand and Misra, Ishan},
  booktitle={CVPR},
  pages={},
  year={2023}
}

@inproceedings{shafir2023human,
  title={Human Motion Diffusion as a Generative Prior},
  author={Shafir, Yonatan and Tevet, Guy and Kapon, Roy and Bermano, Amit H},
  booktitle={ICLR},
  year={2024}
}

@inproceedings{fan2024freemotion,
  title={{FreeMotion}: A Unified Framework for Number-Free Text-to-Motion Synthesis},
  author={Fan, Ke and Tang, Junshu and Cao, Weijian and Yi, Ran and Li, Moran and Gong, Jingyu and Zhang, Jiangning and Wang, Yabiao and Wang, Chengjie and Ma, Lizhuang},
  booktitle={ECCV},
  pages={},
  year={2024},
  organization={Springer}
}

@inproceedings{petrovich21actor,
  title={Action-Conditioned {3D} Human Motion Synthesis with Transformer {VAE}},
  author={Petrovich, Mathis and Black, Michael J. and Varol, G{\"u}l},
  booktitle={ICCV},
  year={2021}
}

@inproceedings{guo2020action2motion,
  title={{Action2Motion}: Conditioned Generation of {3D} Human Motions},
  author={Guo, Chuan and Zuo, Xinxin and Wang, Sen and Zou, Shihao and Sun, Qingyao and Deng, Annan and Gong, Minglun and Cheng, Li},
  booktitle={ACM MM},
  pages={},
  year={2020}
}

@article{heusel2017gans,
  title={GANs Trained by a Two Time-Scale Update Rule Converge to a Local Nash Equilibrium},
  author={Heusel, Martin and Ramsauer, Hubert and Unterthiner, Thomas and Nessler, Bernhard and Hochreiter, Sepp},
  journal={NeurIPS},
  volume={},
  number={},
  pages={},
  year={2017}
}

@inproceedings{onuma2008fmdistance,
  title={{FMDistance}: A Fast and Effective Distance Function for Motion Capture Data},
  author={Onuma, Kensuke and Faloutsos, Christos and Hodgins, Jessica K.},
  booktitle={Eurographics},
  year={2008}
}

@incollection{muller2005efficient,
  title={Efficient Content-Based Retrieval of Motion Capture Data},
  author={M{\"u}ller, Meinard and R{\"o}der, Tido and Clausen, Michael},
  booktitle={SCA},
  year={2005}
}

@inproceedings{song2020denoising,
  title={Denoising Diffusion Implicit Models},
  author={Song, Jiaming and Meng, Chenlin and Ermon, Stefano},
  booktitle={ICLR},
  year={2021}
}

@inproceedings{nichol2021improved,
  title={Improved Denoising Diffusion Probabilistic Models},
  author={Nichol, Alexander Quinn and Dhariwal, Prafulla},
  booktitle={ICML},
  pages={},
  year={2021},
  organization={PMLR}
}

@inproceedings{loshchilov2017decoupled,
  title={Decoupled Weight Decay Regularization},
  author={Loshchilov, Ilya and Hutter, Frank},
  booktitle={ICLR},
  year={2019}
}

@inproceedings{radford2021learning,
  title={Learning Transferable Visual Models From Natural Language Supervision},
  author={Radford, Alec and Kim, Jong Wook and Hallacy, Chris and Ramesh, Aditya and Goh, Gabriel and Agarwal, Sandhini and Sastry, Girish and Askell, Amanda and Mishkin, Pamela and Clark, Jack and Krueger, Gretchen and Sutskever, Ilya},
  booktitle={ICML},
  pages={},
  year={2021}
}

@article{gemma2024report,
  title={Gemma 2: Improving Open Language Models at a Practical Size},
  author={Gemma Team and Mesnard, Thomas and Hardin, Cody and Dadashi, Robert and Bhupatiraju, Surya and Pathak, Shreya and Sifre, Laurent and Riviere, Morgane and Kale, Mihir and Love, Julien and Tafti, Pouya and Hussenot, Leopold and Sessa, Pier Giuseppe and Chowdhery, Aakanksha and others},
  journal={arXiv},
  year={2024}
}

@misc{vicuna2023,
  title={Vicuna: An Open-Source Chatbot Impressing GPT-4 with 90\%* ChatGPT Quality},
  howpublished={https://lmsys.org/blog/2023-03-30-vicuna/},
  author={Chiang, Wei Lin and Li, Zhuohan and Lin, Zi and Sheng, Ying and Wu, Zhanghao and Zhang, Hao and Zheng, Lianmin and Zhuang, Siyuan and Zhuang, Yonghao and Gonzalez, Joseph E. and Stoica, Ion and Xing, Eric P.},
  month={},
  note         = {Accessed: 2025-05-06},
  
  year={2023}
}

@inproceedings{naeem2020reliable,
  title={Reliable Fidelity and Diversity Metrics for Generative Models},
  author={Naeem, Muhammad Ferjad and Oh, Seong Joon and Uh, Youngjung and Choi, Yunjey and Yoo, Jaejun},
  booktitle={ICML},
  pages={},
  year={2020},
  organization={PMLR}
}

@inproceedings{li2022blip,
  title={BLIP: Bootstrapping Language-Image Pre-training for Unified Vision-Language Understanding and Generation},
  author={Li, Junnan and Li, Dongxu and Xiong, Caiming and Hoi, Steven},
  booktitle={ICML},
  year={2022}
}

@inproceedings{zhang2023video,
  title={Video-{LLaMA}: An Instruction-tuned Audio-Visual Language Model for Video Understanding},
  author={Zhang, Hang and Li, Xin and Bing, Lidong},
  booktitle={EMNLP Demo},
  year={2023}
}

@article{wan2025,
  title={Wan: Open and Advanced Large-Scale Video Generative Models},
  author={{Wan Team}},
  journal={arXiv},
  year={2025}
}

@inproceedings{deshmukh2023pengi,
  title={Pengi: An Audio Language Model for Audio Tasks},
  author={Deshmukh, Soham and Elizalde, Benjamin and Singh, Rita and Wang, Huaming},
  booktitle={NeurIPS},
  year={2023}
}

@inproceedings{deng2024musilingo,
  title={MusiLingo: Bridging Music and Text with Pre-trained Language Models for Music Captioning and Query Response},
  author={Deng, Zihao and Ma, Yinghao and Liu, Yudong and Guo, Rongchen and Zhang, Ge and Chen, Wenhu and Huang, Wenhao and Benetos, Emmanouil},
  booktitle={NAACL Findings},
  pages={},
  year={2024}
}

@inproceedings{wu2024motion,
  title={{Motion-Agent}: A Conversational Framework for Human Motion Generation with {LLM}s},
  author={Wu, Qi and Zhao, Yubo and Wang, Yifan and Liu, Xinhang and Tai, Yu-Wing and Tang, Chi-Keung},
  booktitle={ICLR},
  year={2025}
}

@inproceedings{luo2024m3gpt,
  title={{M$^3$GPT}: An Advanced Multimodal, Multitask Framework for Motion Comprehension and Generation},
  author={Luo, Mingshuang and Hou, Ruibing and Li, Zhuo and Chang, Hong and Liu, Zimo and Wang, Yaowei and Shan, Shiguang},
  booktitle={NeurIPS},
  year={2024}
}

@inproceedings{zhang2024finemogen,
  title={{FineMoGen}: Fine-Grained Spatio-Temporal Motion Generation and Editing},
  author={Zhang, Mingyuan and Li, Huirong and Cai, Zhongang and Ren, Jiawei and Yang, Lei and Liu, Ziwei},
  booktitle={NeurIPS},
  year={2023}
}

@inproceedings{zhang2025social,
  title={Social Agent: Mastering Dyadic Nonverbal Behavior Generation via Conversational LLM Agents},
  author={Zhang, Zeyi and Zhou, Yanju and Yao, Heyuan and Ao, Tenglong and Zhan, Xiaohang and Liu, Libin},
  booktitle={SIGGRAPH Asia},
  pages={},
  year={2025}
}

@inproceedings{jiang2025solami,
  title={Solami: Social vision-language-action modeling for immersive interaction with 3d autonomous characters},
  author={Jiang, Jianping and Xiao, Weiye and Lin, Zhengyu and Zhang, Huaizhong and Ren, Tianxiang and Gao, Yang and Lin, Zhiqian and Cai, Zhongang and Yang, Lei and Liu, Ziwei},
  booktitle={CVPR},
  pages={},
  year={2025}
}

@inproceedings{li2021ai,
  title={AI Choreographer: Music Conditioned 3D Dance Generation with {AIST}++},
  author={Li, Ruilong and Yang, Shan and Ross, David A. and Kanazawa, Angjoo},
  booktitle={ICCV},
  pages={},
  year={2021}
}

@inproceedings{li2023finedance,
  title={FineDance: A Fine-grained Choreography Dataset for 3D Full Body Dance Generation},
  author={Li, Ronghui and Zhao, Junfan and Zhang, Yachao and Su, Mingyang and Ren, Zeping and Zhang, Han and Tang, Yansong and Li, Xiu},
  booktitle={ICCV},
  pages={},
  year={2023}
}

@misc{canadasalsacongress_rules_2024,
  title        = {Rules, Definitions and Judging Criteria ~ 2024},
  author       = {{Canada Salsa and Bachata Congress}},
  year         = {2024},
  howpublished          = {https://www.canadasalsacongress.com/rules},
  note         = {Accessed: 2025-05-06}
}

@book{bartneck2024human,
  title={Human-robot interaction: An introduction},
  author={Bartneck, Christoph and Belpaeme, Tony and Eyssel, Friederike and Kanda, Takayuki and Keijsers, Merel and {\v{S}}abanovi{\'c}, Selma},
  year={2024},
  publisher={Cambridge University Press}
}

@inproceedings{zhang2024react,
  title={React to this! how humans challenge interactive agents using nonverbal behaviors},
  author={Zhang, Chuxuan and Burkanova, Bermet and Kim, Lawrence H and Yip, Lauren and Cupcic, Ugo and Lall{\'e}e, St{\'e}phane and Lim, Angelica},
  booktitle={IROS},
  pages={},
  year={2024},
  organization={IEEE}
}

@inproceedings{yang2025omniretarget,
  title={{Omniretarget: Interaction-preserving data generation for humanoid whole-body loco-manipulation and scene interaction}},
  author={L. Yang et al.},
  booktitle={ICRA},
  pages={},
  year={2025},
  organization={IEEE}
}

@article{fitch2016dance,
  title={{\DIFaddFL{Dance, music, meter and groove: a forgotten partnership}}},
  author={\DIFaddFL{Fitch, W Tecumseh}},
  journal={\DIFaddFL{Frontiers in human neuroscience}},
  volume={},
  pages={},
  year={\DIFaddFL{2016}},
  publisher={\DIFaddFL{Frontiers Media SA}}
}

@article{radford2019language,
  title={{{Language Models are Unsupervised Multitask Learners}}},
  author={{A. Radford et al.}},
  journal={{OpenAI Technical Report}},
  year={{2019}}
}

@inproceedings{cao2025real,
  title={{{Motionctrl: A real-time controllable vision-language-motion model}}},
  author={{B. Cao et al.}},
  booktitle={{ICCV}},
  pages={},
  year={{2025}},
  organization={{IEEE}}
}

@article{lee2026implicit,
        title={{{Implicit Maximum Likelihood Estimation for Real-time Generative Model Predictive Control}}},
        author={{{G. Lee et al.}}},
        journal={{{ICRA}}},
        year={{{2026}}},
}

\end{document}